\documentclass[letterpaper, times]{article}

\pdfoutput=1

\usepackage{moreverb,url}

\usepackage[colorlinks,bookmarksopen,bookmarksnumbered,citecolor=red,urlcolor=red]{hyperref}

\newcommand\BibTeX{{\rmfamily B\kern-.05em \textsc{i\kern-.025em b}\kern-.08em
T\kern-.1667em\lower.7ex\hbox{E}\kern-.125emX}}

\usepackage{graphicx} 
\usepackage{tabularx}
\usepackage{amsmath} 
\usepackage{amssymb}  
\usepackage{algorithm}
\usepackage[noend]{algpseudocode}
\usepackage[colorinlistoftodos]{todonotes}
\usepackage{soul,color}
\usepackage{natbib}

\usepackage{url, hyperref, lineno,microtype,subcaption}
\usepackage{csquotes}
\usepackage[onehalfspacing]{setspace}

\graphicspath{ {figures/} }

\DeclareMathOperator{\E}{\mathbb{E}}

\providecommand{\keywords}[1]{\textbf{\textit{Keywords: }} #1}

\begin{document}


\title{Affordance as general value function:\\ A computational model}

\author{Daniel Graves, Johannes G\"unther and Jun Luo}
\date{}



\maketitle

\begin{abstract}
General value functions (GVFs) in the reinforcement learning (RL) literature are long-term predictive summaries of the outcomes of agents following specific policies in the environment.
Affordances as perceived action possibilities with specific valence may be cast into predicted policy-relative goodness and modelled as GVFs.
A systematic explication of this connection shows that GVFs and especially their deep learning embodiments (1) realize affordance prediction as a form of direct perception, (2) illuminate the fundamental connection between action and perception in affordance, and (3) offer a scalable way to learn affordances using RL methods.
Through an extensive review of existing literature on GVF applications and representative affordance research in robotics, we demonstrate that GVFs provide the right framework for learning affordances in real-world applications.
In addition, we highlight a few new avenues of research opened up by the perspective of ``affordance as GVF'', including using GVFs for orchestrating complex behaviors.
\end{abstract}


\section{Introduction}
J. J. Gibson first used the word ``affordance''  in \citep{gibson1966} and explicated the notion in \citep{gibson1979} as follows (p.127):

\begin{displayquote}
The \textit{affordances} of the environment are what it \textit{offers} the animal, what it \textit{provides} or \textit{furnishes}, either for good or ill. ...
I mean by [affordance] something that refers to both the environment and the animal in a way that no existing term does.
It implies the complementarity of the animal and the environment.
\end{displayquote}

A mundane example is how staircases afford walking.
The vertical ``rise'' of a step must not be too big or too small so that a single step is easy to climb and the total number of steps is not too many.
The horizontal ``run'' of a step must not be too narrow or too wide so that one foot fits comfortably while the other foot reaches the next step easily.
Whatever the metric sizes of the rise and run are, what matters for the stair climber is how the staircase affords their walking up and down, given their hip height \citep{warren1984JEPHPP}, stride length \citep{cesari2003}, and leg strength \citep{konczak1992JEPHPP}.
In this sense, affordances are \textit{action possibilities} considered as relations between the embodied agent and the environment concerning specific forms of \textit{fit or misfit}.

In addition, \citet[pp.133-5]{gibson1979} emphasizes that perception of affordances is \textit{direct} in the sense of involving no deliberative reasoning or inference on the basis of geometric measurement or object detection.
Patterns well correlated with the staircase's affordance are readily discernible in the optical array: to see how steep the staircase is from below, one only needs to note how much of the optical array is filled by the rises versus the runs; from above, one only needs to note how much the edges of the steps occlude the runs below.
Crucially, all such perceiving happens from one's own eye height, enabling an automatic ego-relativization that renders unnecessary the mediation of third-person metric measurements of the rises and runs \citep{Mark1987JEPHPP}.
From the bottom of a staircase designed to suit adult body characteristics \citep{warren1984JEPHPP, cesari2003, konczak1992JEPHPP}, an infant sees mostly rises and barely any runs. They are thus unlikely to perceive it as suitable for walking \citep{Mark1987JEPHPP}. Indeed, when infants first ascend stairs independently, more than 90\% opt for crawling \citep{berger2007ibd}.

\begin{figure}
    \centering
    \includegraphics[width=\columnwidth]{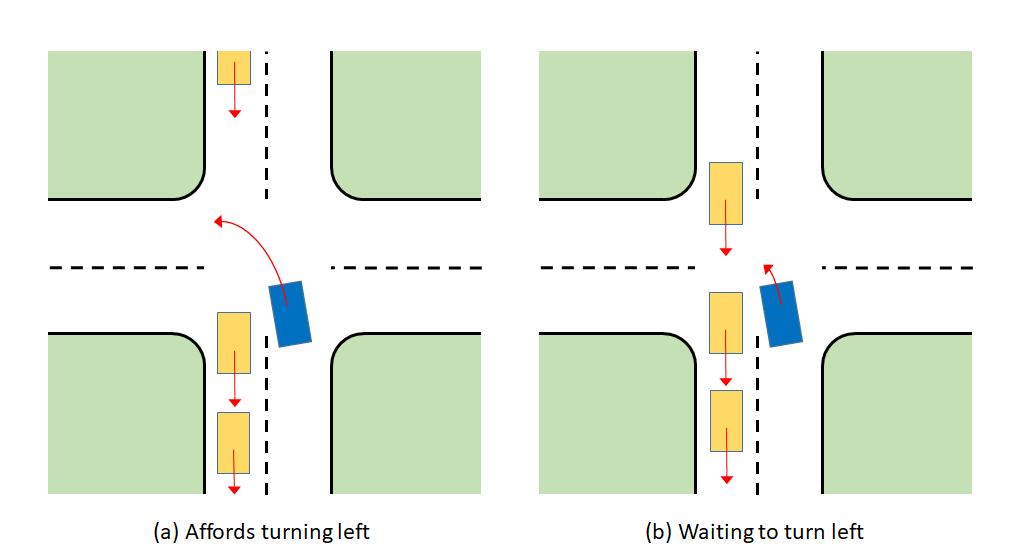}
    \caption{Affordance for unprotected left turns in autonomous driving. (a) The situation affords a safe left turn immediately but not later. (b) The situation affords collision, if the blue vehicle does not wait, and affords successful left turn if it waits a bit.}
    \label{fig_unprotected_left_turn}
\end{figure}

To specifically motivate the relevance of affordance to robotics, let us consider an example from autonomous driving, which is a form of mobile robotics.
The unprotected left turn scenario illustrated in Figure \ref{fig_unprotected_left_turn} is challenging, first of all because the situation is highly dynamic: a second ago, there was no gap, and now there is one; and half a second later, the gap is gone.
Affordance for successful left turn quickly morphs into affordance for collision, without a clear-cut or stable boundary.
Secondly, the situation is highly interactive. Action of the robot vehicle itself affects how affordances unfold. Whereas squeezing into the junction area could engender the gap needed \textit{if} the oncoming vehicles are cooperative enough, merely waiting could encourage them to never slow down.
Finally, given the real-world complexity here, hand-written rules and hand-tuned models are unlikely going to be adequate and learning is necessary \citep{sutton2019bitter}. 

In motor control, there are many situations where basic physics and physiology rule out a sustained control loop. To meet a baseball travelling at 100mph, for example, the batter must start swinging the bat when the ball is still only halfway away. Detailed studies show that instead of constantly adjusting their motion according to the observed trajectory of the fast approaching ball, the batter decides about and initiates the swing based on a predictive reading of the overall situation based on the pitcher's body language and the ball's initial trajectory \citep{gray2002markov, stadler2007baseball, muller2015baseball}. As \citet{epstein2013sports} puts it, ``the only way to hit a ball travelling at high speed is to be able to see into the future.'' 

While forward dynamics models may still be useful to synthesize step-by-step surrogate feedback signals to close the control loop, basic physical constraints of timing and inertia favor direct perception of affordances.
What matters is that such affordances be perceived accurately and quick enough, in spite of the fast-changing contextual factors.
It may be argued that much of the training or learning involved in human sports has to do with learning to perceive affordances \citep{yarrow2009sports, epstein2013sports}.
What these considerations highlight is the fact that affordances are intrinsically \textit{predictive}, but predictive not in the sense of tracking how the system dynamics unfold step by step but rather what the anticipated behavior consequences could be, given the overall situation and the abilities of the agent.

We believe that the perspective of affordance is highly valuable for artificial intelligence (AI) and robotics.
We believe that robotics is the right field for demonstrating its theoretical value for AI in general.
This is only natural because, as illustrated by our examples above, affordance is primarily concerned with action or behavior possibilities by an embodied agent interacting with the physical world. 
We believe that the concept of affordance and the abundant research on it in psychology is a particularly relevant source of inspiration as deep learning (DL) and reinforcement learning (RL) become more and more widely used in robotics.
We believe that a key to unlocking the potential of affordance for robotics in the era of ``second-wave AI'' \citep{smith2019promise} is a suitable computational model of affordance.
In this paper, we set out to propose such a model.

We expect our proposed model to meet a few desiderata. Theoretically, to honor the two key points \citet{gibson1979} emphasized in his original proposal, it should (a) substantiate the directness of affordance perception and (b) explicate the complementarity between action and perception. Practically, to be useful in real-world robotics, it should (c) enable effective learning of affordances from real-world data and (d) scale up to real-world complexity. We call these desiderata, respectively, (a) \textit{directness}, (b) \textit{complementarity}, (c) \textit{learnability}, and (d) \textit{scalability}.

Our specific proposal is that the general value functions (GVFs) framework from RL provides a computational model of affordance that meets these desiderata.
In the rest of the paper, we first motivate this perspective through reviewing some related works.
Next, we present the GVF model of affordance and explain how it meets the desiderata.
We then survey representative real-world applications from both GVF literature and affordance literature to (1) link these two bodies of research under our formalism of affordances as GVFs, and (2) demonstrate the practicality of affordances as GVFs.
Before concluding, we also introduce some important open issues and directions for further investigation.
Our goal is to provide a compelling argument for the theoretical and practical value of the GVF as a computational model of affordance and to highlight how this novel perspective may help deepen collaborative research on affordance in robotics and related fields towards bringing the power of RL to bear on real-world applications through affordances as GVFs.

\section{Related Works}
A large body of empirical and theoretical work followed after Gibson's original proposal. As an early example,
\citet{lee1976ttc} first identified a form of perceptual invariance in car driving that corresponds to time to collision (TTC) and can be used to control braking and stopping.
Much subsequent research highlighted the importance of affordance for understanding performance in sports, as surveyed by \citet{fajen2008sport}.
Eleanor Gibson and collaborators \citep{gibson2000book} turned out a large body of work on learning of affordance, including the famous visual cliff experiments.
And \cite{norman1988} popularized the concept of affordance for interaction design.
More theoretical and philosophical treatment can be found in \citep{turvey1992, chemero2003, noe2004}.
In this section, we can only briefly survey works on computational models of affordance in robotics and works that help us explicate the connection between affordance and RL through the perspective of prediction.

\subsection{Affordance Modeling in Robotics}

Since the seminal work by \citet{Duchon1998ER}, which treated traversability using control laws based on optical flow, there has been a steady stream of research on putting the notion of affordance to good use in robotics.
When it comes to systematizing affordance research for robotics, \citet{sahin2007} proposed one of the earlier computational frameworks. This proposal treats learning of affordances in terms of clustering of features into equivalence classes.

A recent survey of computational models of affordance in robotics by \citet{zech2017} showed that in spite of a myriad of proposals, the notion of affordance does not yet have a computational formulation that both is widely useful in robotics and does justice to the original insights of Gibson.
The authors note in particular a tension between the use of computational representations and Gibson's claim that the perception of affordance is direct.
They also note the widespread use of supervised learning methods and offline learning setups and call for more research into the ecologically more realistic semi-supervised and self-supervised learning strategies, a point echoed by another survey \citep{Yamanobe2017ABR} on affordance for robotic manipulation.
Moreover, \citet{zech2017} highlighted a scarcity in research on how affordances may be chained or hierarchically coordinated for appropriate behavior patterns.
Finally, they emphasized the importance of predictions for planning with affordances.
For our purposes, their comprehensive survey has highlighted four important open issues for computational models of affordance: (a) that of \textit{directness of affordance perception}, (b) that of \textit{learning strategies}, (c) that of \textit{behavior structures based on affordances}, and (d) that of \textit{using affordances in planning and control}.

Further surveys on computational models of affordance have been presented by \citet{thill2013} and \citet{jamone2018}. Similar to \citep{sahin2007} mentioned above, these models tend to center around sensing features or properties of objects and even detecting them through object classifiers in the environment. The four open issues remain to be adequately addressed.

Among publications on affordance in robotics, not too many specifically deal with affordance from an RL perspective.
\cite{Toussaint2003LearningWM} uses the value iteration method of RL as a way to bind a world model to goal-specific predictions about traversability.
\cite{Cruz2016TrainingAW} use affordance specification to identify actions that are possible given a context and thereby trim the action space so as to better focus the RL process on the relevant actions.
\citet{Paletta2007PerceptionAD} and \citet{wu2020spatial} are most explicit about using RL to learn predictive features that underwrite affordance perception.
While these are relevant precursors to our proposal, with \citep{Paletta2007PerceptionAD} and \citep{wu2020spatial} being the closest in terms of general direction, none of them treat affordance itself systematically and rigorously from the RL perspective as we propose to do.

\subsection{Prediction, Affordance, and RL}

Predictions play an essential role in an agent's ability to deal with the world \citep{clark2013, epstein2013sports}. 
While \citet{gibson1979} himself does not explicitly theorize the predictive aspects of affordance, researchers working on perceptual mechanism \citep{lee1976ttc} or computational model of affordance \citep{ugur2009, zech2017} have inevitably noticed the intrinsic temporal aspect of affordance.
For example, collision afforded by the traffic at an intersection is a future consequence of possible actions now (Figure \ref{fig_unprotected_left_turn}).
While the good or ill of an action available now will not fully materialize until later, affordance perceived as such binds them into a single unit of concern: action possibilities available \textit{now} along with their \textit{predicted} valence.

In RL, the value function maps a state to the ``value'', i.e. the expected cumulative reward.
In other words, it makes a prediction about how good the agent's subsequent performance could be, given its current position in the environment, i.e. ``state'', and its specific way of acting, i.e. ``policy'' \citep{sutton1998}.
The agent's objective is to learn a policy that maximizes value.
Simple as this perspective is, it has proved to be extraordinarily powerful when suitable learning mechanisms, such as neural networks, could be used to learn highly complex value functions and policies.
An early illustration is TD gammon \citep{tesauro1995} that rivals human champions in Backgammon.
More recent successes using this approach range from video games \citep{mnih2013, oriol2019starcraft2} to board games \citep{silver2016, silver2018} to physically manipulating Rubik's cube \citep{openai2019rubiks}.

The notion of value in RL is similar to affordance in that it predicts the valence of the policy being learned.
In fact, affordance predictions have been used with great success in the past for a number of real-world problems such as bin picking \citep{zeng2019amazon}, navigation \citep{ugur2007traversability}, and mobile manipulation \citep{wu2020spatial}.
In each of these examples, the final outcome of at least one action possibility is predicted.
The prediction model is often trained by supervised learning \citep{zech2017}, with RL used occasionally, e.g. \citep{wu2020spatial}.
It is important to note, however, that conventional supervised learning of affordance prediction can be cast as a special case of value prediction \citep{sutton2018}.
In other words, while there are important differences in how these algorithms learn, value prediction is a broader concept that encompasses many more applications of affordances in robotics.
Both supervised learning and temporal difference (TD) learning can be used to learn value \citep{sutton1988, sutton2018} with TD learning being the more typical method of learning value with RL.
The differences between learning affordance predictions conventionally with supervised learning, e.g. \citep{ugur2007traversability, zeng2019amazon}, and learning value with temporal difference learning, e.g. \citep{wu2020spatial}, are summarized in Table \ref{tab:prediction_learning_comparison}.


\begin{table}[ht]
    \centering
    \caption{Comparison of two common methods of learning value: Supervised affordance learning vs. Temporal difference learning}
    \begin{tabular}[t]{ |p{3.75cm}|p{3.75cm}| }
        \hline
        \textbf{Supervised Learning} & \textbf{TD Learning} \\
        \hline
        Update based on target label collected much farther into the future & Update based on current cumulant and estimated next prediction \\
        \hline
        Higher memory demand for saving entire trajectories for offline learning & Lower memory demand with online learning \\
        \hline
        Higher variance updates make selection of hyper-parameters difficult & Lower variance updates for better learning stability \\
        \hline
        Unbiased learning & Biased learning during initial stages of learning\footnote{Biased learning of value has a complex impact on learning but the lower variance updates often is more helpful to speed up and improve learning stability compared to supervised learning \citep{sutton1988, sutton2018}.} \\
        \hline
        Difficult to learn off-policy & Can be learned off-policy \\
        \hline
    \end{tabular}\label{tab:prediction_learning_comparison}
\end{table}

The core intuition behind our proposal is that \textit{action-relative valence of affordance could be viewed as a form of prediction similar to value prediction in RL}. In other words, value learning in RL could be productively exploited in a principled way for learning affordance perception. The two most common methods of learning value as summarized in Table \ref{tab:prediction_learning_comparison} each have its advantages and disadvantages. While many applications of affordance use the supervised learning method \citep{ugur2007traversability, zeng2019amazon}, it is actually only a special approach to the general value learning problem, which can also be effectively tackled with methods of RL, especially the TD learning methods.

Now, we immediately face a problem: RL prediction is concerned only with the expected accumulated \textit{reward}, but affordances in the real world are many and highly varied.
While the reward-centered value formulation of RL may be suitable for single-purpose games, predicting and maximizing nothing but value is too restrictive for real-world robotics, where many aspects of the robot-environment relationship will likely need to be specifically attended to.
The Horde architecture \citep{sutton2011} addresses this issue by expanding the conventional interpretation of value functions in RL to \textit{general value functions} (GVFs). 
Whereas a value function predicts about future reward, which is a specific scalar, a general value function predicts about arbitrary scalars \citep{white2017}.
GVFs could be used to evaluate outcomes of policies on any aspect that could be adequately represented as a scalar, such as the robot's sensor data.
Because the GVF prediction is a straightforward generalization of value function prediction, the comparison summarized in Table \ref{tab:prediction_learning_comparison} also applies to GVF learning.

One important feature of GVFs is that they could be learned using existing RL methods, especially TD methods for policy evaluation \citep{sutton1998}, or supervised learning of a target accumulated value \citep{sutton2018} as described in Table \ref{tab:prediction_learning_comparison}.
In addition, the GVF predictions could be learned independently of any reward or goal defined in the environment, which opens the possibility of their general utility.
Furthermore, the Horde architecture \citep{sutton2011}, which allows simultaneous learning of many GVFs, was demonstrated as a scalable way of doing predictive learning through the example of learning to predict sensorimotor data streams underwriting the psychological phenomenon of ``nexting'' \citep{modayil2012, awhite2015}.
Later work also showed that there are substantial benefits from layering GVF predictions to create more complex forms of predictions in partially observable environments \citep{schaul2013, schlegel2018}.

However, in spite of these nice properties of GVFs for making predictions, in spite of the well established mechanisms for learning them, and in spite of some specific examples of how GVF predictions may be used to control real-world robots \citep{edwards2016, gunther2018, graves2019a, graves2020}, there is as yet no general computational framework for using such predictions, such as using them in downstream learning or optimization of policies.


Our main contribution is the formulation of a new computational model of affordance through linking affordances and GVFs systematically.
We hope that this new framework will put what is already being practised in real-world robotics research \citep{zeng2019amazon, ugur2007traversability, wu2020spatial} on better theoretical footing, will help address some key open issues in affordance research in robotics \citep{zech2017}, and will ultimately enable impactful use of GVF predictions as embodiments of affordances in the real world.

\section{Affordances as GVFs}
In this section, we formalize the idea of affordances as GVF-based predictions.
As part of the background assumptions, we view affordances as \textit{action possibilities} with specific \textit{valence} that are available to an embodied agent in the environment.
Valence here reflects the fact that these action possibilities are in varying degrees ``either for good or ill'' \citep{gibson1979}.
Given such an understanding, we make a three-part proposal. First, we (1) model \textit{action possibilities} with the notion of \textit{options} \citep{sutton1999} in the RL literature. Second, we (2) model \textit{valences} of action possibilities with \textit{option-specific GVF-based predictions} also from the RL literature.
Lastly, (1) and (2) in combination give us (3) a model of \textit{affordance perception} as \textit{GVF prediction}.
Under this proposal, we may now use relevant RL methods, especially their deep-learning variants, to learn affordances through learning predictions about the consequences of following specific options.
The desiderata of directness, complementarity, learnability, and scalability, in turn, may be met by appealing to the properties of the option-specific GVF model and its associated RL architecture and DL implementation.

\subsection{RL Preliminaries}

Before we can present our three-part proposal formally, we need to introduce the relevant RL elements it uses.

We follow the standard RL practice to model the agent-environment system as a \textit{Markov Decision Process} (MDP) $M$, which is defined by the tuple $(S, A, P, R, \gamma)$.
$S$ is the set of \textit{states} the system could be in.
$A$ is the set of \textit{actions} the agent could take.
$P$ is the \textit{state transition matrix} given by \textit{transition probabilities} $p(s_{t+1}|s_t,a_t)$.
At each time step $t$, the agent takes an action $a_t \in A$ and the system transitions into the next state $s_{t+1} \in S$ according to $P$ and the agent receives a \textit{reward} $r_t \in R$ in state $s_{t+1}$.
The \textit{discount factor} $\gamma$ governs how reward is accumulated over multiple time steps.

The agent decides its actions by following a \textit{policy} $\pi(a|s)$ that gives the probability of choosing an action $a \in A$ in a given state $s \in S$.
Typically, given an MDP, the goal of RL is to find a policy $\pi(a|s)$ that maximizes the \textit{return} $G_t$ defined as the sum of rewards over consecutive time steps and discounted by $\gamma$:  

\begin{equation}
G_t=\sum_{k=0}^{\infty} {\gamma^{k} r_{t+k+1}}
\label{eq_return}
\end{equation}

\noindent
Often this is accomplished by first learning a \textit{value function} that predicts the expected return by following the policy $\pi$:
 
\begin{equation}
V^{\pi}(s)=\E_{\pi}[G_t|s_t=s]
\label{eq_value}
\end{equation}

\noindent
and then deriving a new policy that is based on $V^{\pi}$ but improves over it where $\E_{\pi}$ is the expectation over the state visitation distribution induced by policy $\pi$.
The process of learning $V^{\pi}$ is usually called \textit{policy evaluation} and the process of improving over $V^{\pi}$ is called \textit{policy improvement}. For our purposes, policy evaluation is the focus, as it is the policy evaluation methods in RL that will be deployed to learn affordances.

Consider as an example AlphaGo \citep{silver2016}, which uses RL to train an agent to master the game of Go.
The value function of AlphaGo assesses a state of the game by predicting the win or loss of the agent---the valence of that state for AlphaGo under its policy.
The goal of AlphaGo training is to find a policy that maximizes the predicted valence and win the game. The purpose of the value function is to predict that valence accurately so as to facilitate the maximization.
AlphaGo's value prediction is the \textit{expected} win or loss associated with the agent following its policy, an expectation that summarizes---theoretically speaking---the impact of all possible moves and countermoves according to their probability.
In this sense, we may say that $V^{\pi}$ \textit{synoptically} predicts how well an agent is expected to do by starting from $s$ and following $\pi$.

It is important to note that the definition of the policy $\pi$ above takes a global perspective on the entire MDP, covering the whole state space $S$.
For complex agent-environment systems, such a monolithic decision-making structure is often too rigid and unwieldy, not suitable for flexible reuse of skills that support organization of complex behavior.
To address such limitations, \citet{sutton1999} proposed the notion of \textit{options}.
Intuitively, an option is a specific way of behaving under specific circumstances, or a specific skill suitable for some specific situations.
Formally, an option is the tuple $(I, \beta, \tau)$, where $I \subset S$ is the \textit{initiation set} that contains the states in which the option can be initiated, $\tau$ is the policy of the option, and $\beta$ is the \textit{termination function} of the option.
Besides the fact that its use is subject to the initiation set and termination function of the option, $\tau(a|s)$ is not formally different from $\pi(a|s)$.
The termination function $\beta(s)$ gives the probability for the agent to stop following $\tau$ in state $s$ versus continuing with it.

\subsection{Model Action Possibilities as Options}
The first part of our three-part proposal is to model action possibilities as options in RL.
While this is not a particularly novel proposal \citep{sutton1999}, it is important for reaching the thesis that affordances can be modeled as GVFs.
Specifically, we model an action possibility as an option given by $(I, \beta, \tau)$ where $I$ is the set of states where the action possibility can be executed, $\beta(s)$ is a probability function that determines when the action possibility is completed or terminated, and $\tau$ is the policy of the action possibility defining the action the agent should take in each consecutive state until termination.

What matters about the options formulation is that it brings granularity to the behavior organization of an agent.
On the one hand, options are conceived at the level of \textit{policies} above the lower level of \textit{primitive actions} of the underlying MDP.
Unlike an MDP action, which is neither good or bad in itself, an option is an appropriate unit subject to policy evaluation, with the policy $\tau$ being its core constituent.
That is to say, options have future-oriented good or ill fit with specific situations and therefore can be bearers of specific valence.
On the other hand, the effective scope of the policy of an option is delineated by the initiation set $I$ and the termination function $\beta(s)$.
This way, rather than relying on a monolithic, cover-all policy, an agent's overall competence may be decomposed into many well-orchestrated options.
For these reasons, we take options, rather than either the primitive MDP actions or the monolithic policy, to be a suitable model of the \textit{action possibilities} associated with affordances.

The value of an action possibility of a task defined by a specific reward function is the cumulative sum of rewards collected when following $\tau(a|s)$ until termination according to $\beta(s)$, ignoring any actions the agent may take after terminating the option.
In order to learn the value of an option, we propose a more general form of the value function that takes into account the termination probability:

\begin{equation}
G^{o}_t=\sum_{k=0}^{\infty}{ \left(\prod_{i=0}^{k-1}{(1-\beta(s_{t+i+1}))} \right) \gamma^k  r_{t+k+1}}
\label{eq_option_return}
\end{equation}

\begin{equation}
V^{\tau}_{o}(s)=\E_{\tau}[G^{o}_t|s_t=s]
\label{eq_option_value}
\end{equation}

\subsection{Model Valence Perception as GVF Prediction}

We now turn to the question of how to model valence perception computationally as the second part of our three-part proposal.
We achieve this be generalizing \eqref{eq_option_return} further by replacing the reward $r_t$ with specific concerns called \textit{cumulants} applied across a broad range of possible tasks.

While value function prediction has many properties agreeable to affordance, it is narrowly concerned about accumulated reward only of a specific task.
As noted earlier, the fit between an agent and the environment has many aspects that may all matter simultaneously but in different ways.
Arriving at the destination in time safely is just one of the many concerns an autonomous car is subject to.
Passenger comfort, fuel efficiency, and law-abidingness along the way are important too.
Also worth attending to on a smaller scale are concerns such as ``how likely am I going to be rear-ended if I brake this much?'' and ``how much is the car going to jump when it hit the upcoming speed bump?''
Moreover, predictions are needed not only about many aspects of the agent-environment fit, but also under \textit{many more policies} other than the one directly optimized in RL.
Such predictions about other policies could be useful for general planning purposes as well as in the RL optimization process itself, because they could help answer ``what-if'' questions about the various options the agent is weighing.
This leads to general value functions \citep{sutton2011}.

\textbf{General Value Functions:}
A GVF may be specified by the tuple $(c, \tau, \gamma)$, where
\begin{itemize}
\item $c$, called the \textit{cumulant}, is a scalar-valued step-specific signal from the environment that replaces the reward $r$, 
\item $\tau$ may be \textit{any policy} that can be subject to policy evaluation with respect to $c$, and
\item $\gamma(s)$ is the \textit{continuation function} that generalizes the discount constant to a function of state.
\end{itemize}

\noindent
Under this specification, the generalized return and its corresponding GVF is formally defined as

\begin{equation}
G^{c}_t=\sum_{k=0}^{\infty}{ \left( \prod_{i=0}^{k-1}{\gamma(s_{t+i+1})} \right) c_{t+k+1}}
\label{eq_gvf_return}
\end{equation}

\begin{equation}
V^{\tau}_{c}(s)=\E_{\tau}[G^{c}_t|s_t=s]
\label{eq_gvf}
\end{equation}

\noindent
In the so-called ``control'' setting, where policy is being improved to maximize reward-based return, $c = r$ and $\tau = \pi$; thus Equation \eqref{eq_gvf} specializes into Equation \ref{eq_value}.
Here we focus on the non-control setting, where we are only concerned with the correspondingly generalized form of policy evaluation, where $c$ and $\tau$ may be anything relevant.
Where context warrants it, the subscript and superscript $c$ may be dropped from our presentation below to simplify notation.

For the following reasons, the above generalization allows for a wider range of predictions.
First, the cumulant relates to the valence of a policy, as a discounted sum of future cumulants, in the same way that reward relates to value.
Since any meaningful scalar-valued signal could be used as cumulant, all sorts of valence predictions can be made.
In practice, the cumulant can be hand-engineered features such as safety \citep{graves2019a}, comfort, centeredness \citep{graves2020}, and duration \citep{sutton2011}.
The cumulant can also represent outcomes such as traversability \citep{ugur2007traversability}, success or failure \citep{zeng2019amazon} or human-provided quality labels \citep{gunther2016}.
Second, the generalization of the discount factor to be described by a continuation function allows for the possibility of episodic predictions, or predictions about policies that terminate in a finite amount of time in the environment and thus allow another policy to be executed afterwards \citep{sutton2011}.
An example of an episodic prediction is in \citet{sutton2011} where the valence of the stopping policy is the time for the agent to stop.
Thus, the GVF framework allows prediction of valence for episodic as well as non-episodic policies.
Third, the valence prediction does not need to be directly related to the policy $\pi$ that is expected to maximize the cumulated reward, but could be for any relevant policy $\tau$.
 
\textbf{General Action-Value Functions:}
Similar to how value functions can be extended to \textit{action-value functions} that are functions of both states and actions \citep{sutton1998}, GVFs can also be extended to \textit{general action-value functions} (GAVFs).
The GVF and GAVF are related by expectation over actions, namely $V^{\tau}(s)=\E_{a' \sim \tau}[Q^{\tau}(s,a')]$, and GAVFs are formally defined as

\begin{equation}
\begin{split}
\resizebox{1.0\hsize}{!}{$Q^{\tau}(s,a) = \E_{\tau} \left[ \sum_{k=0}^{\infty} {\left( \prod_{i=0}^{k-1}{\gamma(s_{t+i+1})} \right) c_{t+k+1}}|s_t=s,a_t=a \right]$}
\end{split}
\label{eq_gavf}
\end{equation}

Alternatively, the GAVF $Q^{\tau}(s,a)$ can also be expressed as 

\begin{equation}
\resizebox{.88\hsize}{!}{$Q^{\tau}(s,a) = \E_{\tau}[c_{t+1} + \gamma(s_{t+1}) V^{\tau}(s_{t+1})|s_t=s,a_t=a]$}
\label{eq_gavf_gvf}
\end{equation}

\noindent
In other words, the GAVF predicts the expected discounted sum of cumulants of taking action $a$ in state $s$, receiving cumulant $c$ and then following the policy $\tau$ thereafter.
By tying $Q^{\tau}(s,a)$ explicitly to $V^{\tau}(s)$, this relationship makes it clear that GAVF may be used to obtain a specific sense of how well each of the next available action could fare, not only in the basic sense of leading to a step-specific cumulant $c$, but also in the sense of landing in a future state with a definite general value under $\tau$.
This property of GAVF may be exploited for planning purposes \citep{graves2019a}.

\subsection{Model Affordance Perception as Option-Specific GVF Prediction}
Given the above treatment of action possibilities as options and valence perception as GVF prediction, we are ready to state our model of affordance.
Formally, we define the affordance of a state $s$ as the tuple $(V^\tau_c(s), (I, \beta, \tau))$ where the GVF $V^\tau_c(s)$ conforms to a special class of GVFs where the continuation function $\gamma(s)$ is defined according to the termination function probability $\beta(s)$ of the option

\begin{equation}
\gamma(s)=\gamma (1 - \beta(s))
\label{eq_continuation_function}
\end{equation}

\noindent
where $\gamma$ is the constant discount factor of the MDP.
In other words, $\gamma(s)$ subsumes both the idea of probabilistic continuation (or termination) and the idea of step-wise discount.
Similarly, we define the affordance of an action $a$ in a state $s$ as the tuple $(Q^\tau_c(s, a), (I, \beta, \tau))$.

The full model is presented in Table \ref{tab:affordance}, which shows how various aspects of affordance are specifically modelled.

\begin{table}[ht]
    \centering
    \caption{Affordance Modelled as GVF}
    \begin{tabular}[t]{ |p{4.75cm}|p{2.75cm}| }
        \hline
        \textbf{Aspect of Affordance} & \textbf{Model} \\
        \hline
        State of agent-environment system & $s \in S$ \\
        \hline
        Primitive action available to agent & $a \in A$ \\
        \hline
        Action possibility as skill & $(I, \beta, \tau)$ \\
        \hline
        Any specific concern & Cumulant $c$ \\
        \hline
        Valence of state & GVF $V^\tau_c(s)$ \\
        \hline
        Valence of an action in a state & GAVF $Q^\tau_c(s, a)$ \\
        \hline
        Affordance of a state & $(V^\tau_c(s), (I,\beta, \tau))$ \\
        \hline
        Affordance of an action in a state & $(Q^\tau_c(s, a), (I,\beta, \tau))$ \\
        \hline
    \end{tabular}\label{tab:affordance}
\end{table}

The key move here is that we link the GVF $V^\tau_c(s)$ and the GAVF $Q^\tau_c(s, a)$ with the option $(I, \beta, \tau)$ by learning the GVF according to the option policy $\tau$ and defining the continuation function $\gamma(s)$ according to the termination function $\beta(s)$ of the option and the discount factor $\gamma$, i.e. $\gamma(s)=\gamma(1-\beta(s))$.
The GVF prediction is thus directly coupled to the option's policy and termination function.
Note that formally everything said in the previous section, about how $V^\tau_c$ and $Q^\tau_c$ are related to a general $\tau$ and a general $\gamma$, applies to our proposal as well.
This is why the linking them here is theoretically unproblematic.

With such a link in place, the valence predictions are now specific to the options that model action possibilities with their situational appropriateness condition.
In this way, the valence predictions may be focused on only the currently relevant or available options, giving substance to the context-specific action possibility aspect of the notion of affordance.
Note that the initiation set $I$ of the option determines when an option or its underlying policy can or cannot be used.
It effectively acts as an attention or relevance-filtering mechanism on all behaviors the agent is capable of or skills the agent possesses.
Practically, the initiation set can help improve efficiency in learning and planning with options in complex problems, because options can be ignored where they are deemed ineffective or impossible \citep{khetarpal2020}.
More generally, tying valence prediction to options also permits research on option discovery \citep{gregor2016, thomas2017, achiam2018, thomas2018, harutyunyan2019} to be applicable to the discovery of GVFs and concomitantly affordances.

Another important point is that while $V^\tau_c$ and $Q^\tau_c$ are functions of state and action, they are relativized to both the cumulant $c$ and the policy $\tau$.
Formally, $c$ may be any scalar and $\tau$ may be the policy of any option.
Substantively, cumulants should express concerns that are ecologically specific forms of good or ill, e.g. safety, comfort, efficiency, cost, timing, bumpiness of a ride, firmness of a grasp, range remaining etc., about the action possibilities or ``options'' relevant and available in the situation.
In practice, multiple such concerns may simultaneously matter about an option and multiple options may be simultaneously evaluated under the same concern.
These can be easily accommodated by introducing the corresponding cumulants and options and, where appropriate, allow the various $V$s and $Q$s to share underlying computational implementation \citep{sherstan2014,graves2020,ugur2007traversability}.

The theoretical formalism for GVF and GAVF (Equations \ref{eq_gvf} and \ref{eq_gavf}) maps directly from states to generalized values or valences.
As noted above, in practice, the implementations of GVFs and GAVFs often take sensor readings or other forms of observations and yield a valence prediction.
Given that such a prediction is grounded through sensory connections to the agent-environment system, it may also be properly regarded as a form of predictive perception, or ``perception as prediction.''\footnote{A term due to Richard Sutton \& Patrick Pilarski (personal communication).}

The key message is that valence of action possibilities, and thus affordance, can be modelled as \textit{option-specific GVF prediction} with the full expressiveness of the model structures summarized in Table \ref{tab:affordance}.
Following existing practice in RL, these model structures can be readily translated into their computational implementations.
We will now turn to methods for computationally learning $V^\tau_c$ and $Q^\tau_c$, which will complete our proposed computational model of affordance.

\subsection{Learning to Perceive Affordances}
There are primarily two ways to learn GVF predictions \citep{sutton2018}.
The first approach involves using supervised learning to predict the valence.
This approach is commonly done in the affordance literature \citep{zeng2019amazon,wu2020spatial, ugur2007traversability} although rarely done in the RL literature \citep{sutton2018}.
The second approach of using temporal difference learning is more common in the RL literature \citep{silver2016,mnih2013,wu2020spatial,sutton1988,sutton2018} as a very general method for policy evaluation through learning expectations of the form in Equation \ref{eq_value}. 
The reasons for the dominance of temporal difference (TD) learning methods in RL is summarized in Table \ref{tab:prediction_learning_comparison}.  
It is generally more efficient due to online learning\footnote{Recent advances in deep learning require a replay buffer which dramatically increases memory requirements in learning}, (2) more stable learning through lower variance updates, and (3) more efficient at learning from samples collected under other policies (i.e. off-policy learning).
We leave a more detailed comparison of the two methods along with a few algorithms for learning GVFs in the Appendix.

The improved efficiency is a key advantage in learning GVFs since it is possible to learn GVFs when the \textit{target policy} $\tau$, i.e. the policy of the GVF that is being learned, is different from the \textit{behavior policy} $\mu$, i.e. the policy that is used to collect the data \citep{sutton2011}.
This is commonly referred to as \textit{off-policy learning} \citep{sutton2018}.
This allows the agent to improve multiple GVF predictions from the same data being collected.
Learning from off-policy experience does present some challenges as the variance of the updates can increase substantially if the behavior policy distribution $\mu(a|s)$ and target policy distribution $\tau(a|s)$ are very different from each other.
\citet{schlegel2019} studied this problem and proposed a way to improve learning by reducing the variance of the updates.
While it does not address the problem when the distributions are vastly different from each other, it does improve learning.

In the supervised learning case, the agent's behavior policy is usually the same as the prediction policy $\tau$.
The reason is that one must compute the importance sampling ratio between the probability of trajectories which suffers high variance possibly infinite variance \citep{sutton2018} which is extremely challenging to learn especially when the trajectories are long.
Thus, the usual way to learn predictions with supervised learning as done in \citep{zeng2019amazon} is to choose random initial states and run the policy $\tau$ to completion where a success or fail label is received to training the prediction.
Unfortunately, this is very data inefficient as the agent can only use data collected under $\tau$ to learn predictions on the outcomes of $\tau$.

\subsection{Meeting the Desiderata}

We are now in a position to assess how our proposed computational model fares with regard to our desiderata.

Regarding \textit{directness}, insofar as the GVF can be realized as a direct mapping from sensory inputs to valence prediction, the proposed model accounts for the directness of affordance perception very well.
While affordance perception as GVF prediction may take geometric measurements and object detection results as input \citep{graves2019a}, dependency on these are strictly optional. Instead predictions can be made directly from raw sensory inputs \citep{gunther2016, zeng2019amazon, graves2020}.
In contrast, if object detection is a necessary dependency of affordance prediction \citep{hassanin2018}, errors may be introduced through the data labelling needed for setting up the required supervised learning.
More importantly, even when geometric measurements and object detection results are used, which is sometime necessary when for example the affordance in question concerns a specifically individuated and tracked object of interest, such as the cars ahead of and behind the ego car \citep{graves2019a}, no explicit, multi-step reasoning about the measurement readings or detected objects is needed for the prediction to be made.

Moreover, as noted earlier, affordance perception under the GVF model is synoptic.
By predicting the sum of cumulants, as in Equation \eqref{eq_gvf_return}, it is not necessary to know precisely when in the future something ecologically significant will happen but that it will happen.
Computationally, long-term outcomes are being perceived \textit{in constant time}, e.g. through a single forward pass of the neural networks implementing the V or Q predictions \citep{mnih2013, silver2016, gunther2016, graves2019a}, regardless of the complexity of the behavior context.
In this way, it contrasts sharply with methods that unroll a next-step model to predict the long-term outcomes, where iterating an imperfect next-step model leads to compounding errors.  See Table \ref{tab:prediction_next_step_comparison} for a comparison of synoptic (e.g. RL value prediction) and convention model-based predictions of the future.

\begin{table}[ht]
    \centering
    \caption{Conventional Prediction vs. Synoptic Prediction}
    \begin{tabular}[t]{ |p{3.75cm}|p{3.75cm}| }
        \hline
        \textbf{Conventional Prediction} & \textbf{Synoptic Prediction} \\
        \hline
        Single step & Prediction over a period \\
        \hline
        Precise next states & Abstraction over all states \\
        \hline
        Number of iterations linear with length of future period & Constant time regardless of length of future period \\
        \hline
        Compounding error makes long-term prediction hard & Error does not compound; long-term prediction easy \\
        \hline
    \end{tabular}\label{tab:prediction_next_step_comparison}
\end{table}

In practice, such synoptic predictions can be quite useful.
Consider the prediction at a traffic intersection that there will be a collision at some point in the future if the agent follows policy $\tau_1$ but not if it follows policy $\tau_2$.
Precisely when the collision will happen may not be important.
What matters is that one option is more likely going to lead to a collision than the other.
And that is enough for informing the agent about the good or ill of the action possibilities, very much in the spirit of Gibson's conception of direct perception.

It must be made clear that our computational model falls far short of a full account of direct perception. We believe that an adequate account will need to take into consideration the ``personal'' vs. ``sub-personal'' levels of psychological explanation and content ascription \citep{mcdowell1994, noe2004} in a way that honors Gibson's commitment to the so-called ``direct realism'' \citep{gibson1967}. To adequately unpack the ``Gibsonian'' conception of direct perception and direct realism is certainly beyond what we can afford here. For now, we can note that (1) as practicing AI and robotics researchers we should not shy away from taking the elements of our proposed model to be ``representational'' in a sub-personal but nevertheless constitutive and robust sense and (2) the specific model we propose operates at the sub-personal level in a way that is friendly to directness at the personal level insofar as explicit reasoning is not necessary and direct use of sensory input is possible. In other words, our model supplies a piece of synchronic or ``single-step'', sub-personal mechanism that could play a central role in the agent's ability to synoptically perceive the action possibilities the environment affords the agent at the personal level. If our proposal serves to demonstrate specifically how direct perception may be achieved through well-motivated learning mechanisms, that already gives a lot of principled support to the direct perception thesis. \footnote{In addition to perception, the issue of directness  is also intertwined with ontology. While we fully agree that agent and environment must be treated together as a whole system ontologically, we disagree with the dynamic systems re-interpretation of Gibson \citep{chemero2007}, which seems to reactionarily concede too much intellectual territory to the simplistic Fodorian conception of representation. Instead, our sense is that an adequate reconstruction of the Gibsonian ``direct realism'' or ``naive realism'' will need to admit a form of ontological pluralism \citep{smith1996o3} that makes sense of how the world \textit{is} for the agent under its own ``natural attitude'', a Husserlian term that was used by \citet[p.~170]{gibson1950} at least once, in contrast to how the world is for the theorist studying the agent.}

When it comes to \textit{complementarity}, it should be amply clear that insofar as valence prediction under our model is always specific to an option, affordances are inseparable from and specific to the agent's skills or abilities.
The complementarity is literally visible in our notation wherein both the agent policy $\tau$ and the environment state $s$ appear in $V^\tau_c(s)$ and $Q^\tau_c(s, a)$.
How the complementarity is captured here, we believe, also gives technical substance to the ``enactive approach to perception'' \citep{noe2004}, according to which an agent's perceptual content is infused with concerns for activity.

Given that our account is based on RL literature and borrows the TD learning methods from there, the desideratum of \textit{learnability} is readily met insofar as valence prediction is concerned.
It is also worth noting that due to the direct perception architecture, affordance perception with GVFs can be learned directly from high dimensional raw sensor data streams, meaning that object detection and thus the effort for data labeling are optional.

However, learnability of the full model as summarized in Table \ref{tab:affordance} is a much bigger challenge.
In fact, how to learn the various components of our proposed model is ongoing active research in the RL community itself.
For example, learning to turn raw input, the Markovian property of which is usually questionable,  into more abstract representations of the state $s \in S$ where the Markovian property could actually hold, is very actively pursued by research on \textit{representation learning} for RL \citep{schaul2013}.
Also, learning the temporal structure in the form of the continuation function $\gamma$ of options so as to allow the predictions to better match the temporal profile of the dynamics of the agent-environment system is an important part of research on \textit{option discovery}.
And, learning the cumulant $c$, which generalizes the learning of the reward $r$ that is traditionally dealt with through \textit{inverse reinforcement learning}, is an open research question that we will return to later.
For now, it suffices to say that our proposal offers a theoretically principled account of how to learn valence prediction for affordance and opens up important new fronts regarding affordance learning by systematically relating it to reinforcement learning.

Finally, our proposed model is \textit{scalable} computationally for the following reasons.
First, GVF predictions may be learned in parallel \citep{sutton2011, modayil2012}, which allows linear scaling in the number of predictions.
We see evidence of this in learning affordances that relate to success/failure of bin picking \citep{zeng2019amazon}, traversability \citep{ugur2007traversability}, and success/failure of mobile manipulation \citep{wu2020spatial} where thousands of predictions structured spatially in a grid pattern are learned.
Second, when combined with deep learning, multiple predictions can be made with the same neural network by sharing earlier layers across different prediction heads, allowing for potentially sublinear scaling in the number of predictions \citep{sherstan2014, graves2020, zeng2019amazon}.
Third, as noted above, for a single prediction, the direct perception architecture operates in constant time regardless of how distant a future the prediction needs to be concerned about.
While this advantage is nothing new in affordance literature \citep{zeng2019amazon}, it is worth pointing out that predicting the future has a long history where conventional approaches of predicting future valence, such as in model-based control, usually can be very difficult to scale computationally since they rely on model-based or sample-based optimization with multi-step roll-outs \citep{corriou2004mpc}.

We may conclude that the affordance as GVF model does an excellent job of meeting the desiderata, but also raises important open questions especially concerning learning the model structures themselves.
\section{Bridging GVF and Affordance Applications}
GVFs have been used in several real-world domains to date, including (1) robots \citep{sutton2011, modayil2012}, (2) prosthetic arms \citep{pilarski2013, edwards2016}, (3) aiding mammals with partial paralysis \citep{dalrymple2020pavlovian}, (4) industrial automation \citep{gunther2016} and (5) autonomous driving \citep{graves2019a, graves2020}. However, none of these examples relate their predictions back to affordances.  Furthermore, a number of real-world applications of affordance prediction can be formalized as GVF predictions; these include (1) traversability \citep{ugur2007traversability}, (2) graspability \citep{zeng2019amazon}, and (3) pushability \citep{wu2020spatial}.  To strengthen our claim that GVFs are a good computational model of affordance, we will discuss how applications of GVFs can be tied to affordances and, in the reverse direction, how several independently developed applications of affordances can be formulated as GVFs.

\subsection{GVF Applications in Light of Affordances}
\noindent
\textbf{Robotics:} 
One of the first applications of GVFs is robotics.
\citet{sutton2011} used GVFs to learn predictions about sensor readings in a mobile robotic platform. The predictions anticipate signals of interest, based on current sensor readings.
Such signals of interest include (1) time before hitting obstacles, and (2) time to stopping on different types of floor materials.
These predictions are made under the assumption of following a fixed policy, tying them to behaviour.
Translated into affordances, these GVFs are used to predict the outcome of following an action---in other words, they predict the affordance of taking said action, if the predicted signal is beneficial.
As shown by \citet{modayil2012}, these predictions can be made in large numbers and for different timescales, closely relating them to \textit{nexting}, which is the phenomenon of a person simultaneously making a large number of predictions about general environment conditions on a short timescale.
While nexting functions at a level that is more fine-grained than how affordance is normally conceived, its successful treatment using the general form of GVF predictions is intriguing. This suggests that when the behavior context warrants highly specific perceptual contingencies be considered at a higher level of ecological interest, such as in \citep{sutton2011}, we may not have to drastically change the underlying mechanism to model them.

\citet{abeyruwan2014} describe an interesting use of GVFs for selecting role assignments of a team of agents in the RoboCup 3D soccer simulation.
An agent can choose one of a fixed number of 13 ``roles'' on the field.
It must predict the value and uniqueness of its role among all other agents without communicating its role to the other agents on the team.
This prediction of uniqueness and value may be interpreted as a kind of affordance prediction to encourage coordinated team play without verbal communication in the simulated soccer game.


\noindent
\textbf{Prosthetic Arms:} 
Recent research has shown the benefit of predictions when used for prosthetic limbs \citep{pilarski2013, sherstan2014, gunther2018, gunther2020examining}, with
\citet{sherstan2014} demonstrating that an abundant number of GVFs can be learned and maintained with reasonable computational resources, allowing these techniques to be used in mobile robots.
Such GVF predictions have been demonstrated to be useful for prosthetic limbs in several ways.
First, they provide an additional source of useful information \citep{sherstan2014, gunther2018}.
Second, they allow better communication between the prosthesis and its wearer \citep{edwards2016}.
From the affordance perspective, these advantages come from the fact that the GVF predictions are learned in an ecological context and through an interactive process wherein the devices are actively engaged. 

\begin{figure}[!t]
\centering
\includegraphics[width = 0.9\columnwidth]{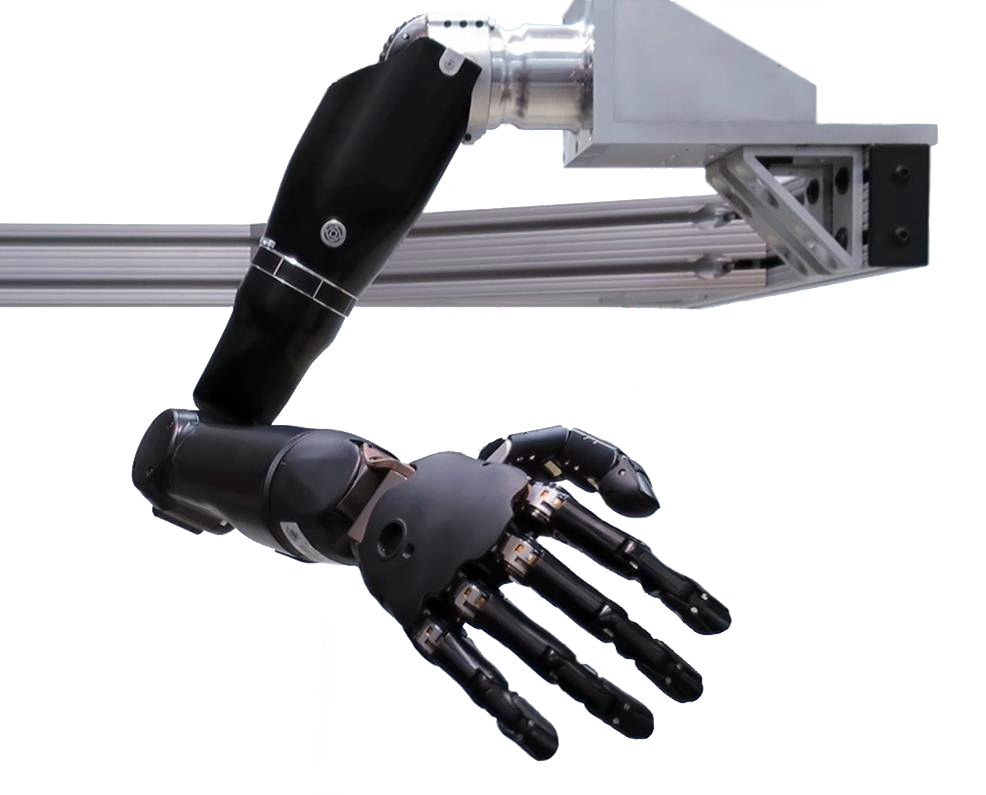}
\caption{The Modular Prosthetic Limb (MPL), a robot arm with many degrees of freedom and sensors used for research in prosthetic limbs \citep{gunther2020examining}.}
\label{fig:mpl}
\end{figure}

Moreover, \citet{gunther2018} have investigated the relationship of errors that occur during learning from surprising external signals.
They use the GVF-based \textit{unexpected demon error} (UDE) \citep{awhite2015} as a measure of surprise.
While the relationship between surprise and affordance is still ongoing research, one can imagine how sensory information that triggers surprise may be directly related to action possibilities.
\citet{jensen2016affect} suggest that emotions have an effect on action-perception.
Thus, one's state of surprise may be viewed as an indicator of one's implicit assumptions about currently available action possibilities that are violated in new context.
An example of this could be an unfamiliar movement with a prosthetic arm that occurs while learning a new task.
Such a surprise informs the learning agent about the need for learning, which can be accommodated by increasing the learning rate.
Another example could be active exploration that is triggered by surprise.
A prosthetic arm that is surprised about its failing to grasp an object would actively explore the shape of the unfamiliar object, thus learning its affordances and improving the handling of it.

\noindent
\textbf{Aiding Mammals with Partial Paralysis: } \citet{dalrymple2020pavlovian} used GVFs in a very unique way to directly interface with living mammals by directly controlling a limb.
They developed a \textit{Pavlovian Control} strategy that utilizes GVF predictions to enable cats with hemisection spinal cord injury to walk. As the intact left hind-limb engages a movement, the GVFs provide the predictions for determining the stimulation needed for the paralyzed right hind-limb to complete the movement pattern, in a way that is similar to a conditioned response to an anticipation.

\citet{dalrymple2020pavlovian} found that the GVFs learned the walking pattern in no more than four steps.
Moreover, in contrast to a purely reaction-based control, the Pavlovian control using GVF predictions requires no manual tuning of threshold, generalizes better across experiment settings, and is capable of recovering from mistakes during walking.
We believe that the advantage of the Pavlovian control here is attributable to the fact that the GVF predictions have captured affordances of the cat's own body.

\noindent
\textbf{Industrial Automation:}
GVFs have been successfully applied in industrial control problems that are too complex to be sufficiently measured or observed and thus have been historically modelled as open-loop processes. Using machine learning---GVFs in particular---these open-loop processes can potentially be transformed into closed-loop processes, allowing them to run reliably without the need for human supervision.

An example in question is industrial laser welding.
Due to lack of appropriate sensors, the quality of welding cannot be measured while the process is running, resulting in the need for time- and labour-intensive quality control. \citet{gunther2016} demonstrate how GVFs can be used to predict the quality of the welding seam from a camera using a combination of deep learning and GVFs. This feedback signal can then be used to inform control decisions, forming a closed control loop. Furthermore, the GVFs enable anticipation of changes in the welding seam quality before they occurred \citep{gunther2018machine}, allowing for predictive control.

The effective use of GVF predictions---predicted weld quality in this example---to support the construction and learning of effective control policies supports our general thesis that GVF predictions represent affordances---quality for the action of welding here in this example.

\noindent
\textbf{Autonomous Driving:}
Autonomous driving is a growing field of research for deep reinforcement learning that tests the limits of modern robotics applications.
Recently, GVFs have been demonstrated to be effective in adaptive cruise control \citep{graves2019a} and vision-based steering \citep{graves2020}.
These works highlight the promise for broader application of GVF predictions to autonomous driving.

\citet{graves2019a} trained GAVFs that predicted future safety, speed, and closeness to the desired target speed for a fixed target policy $\tau$.
The GAVFs, which are a function of state $s$ and action $a$, were used as a partial model to predict the desired outcomes of safety and speed when following the target policy beginning with action $a$.
For example, the safety prediction answered the following question: ``what will my safety be if I take action $a$ and follow similar actions according to the target policy there after.''
These kinds of predictive questions are affordances that permit an agent to plan by inferring a good policy from a limited number of hypothetical ``what-if'' predictions.

\citet{graves2019a} showed that a limited number of predictions could be used to control both a real robot and an autonomous vehicle from a high dimensional LIDAR sensor data stream with performance competitive with existing solutions.

This work was extended to the more complex task of driving a vehicle from vision-based sensors \citep{graves2020}.  GVFs were used to steer a vehicle with just a front-facing camera by learning a limited number of predictions of the future lane centeredness and angle offset with the road over multiple time horizons. Not only was the robot able to steer well on unseen roads but it was also able to slow down for tight corners.

\begin{figure}
    \centering
    \includegraphics[width=0.45\textwidth]{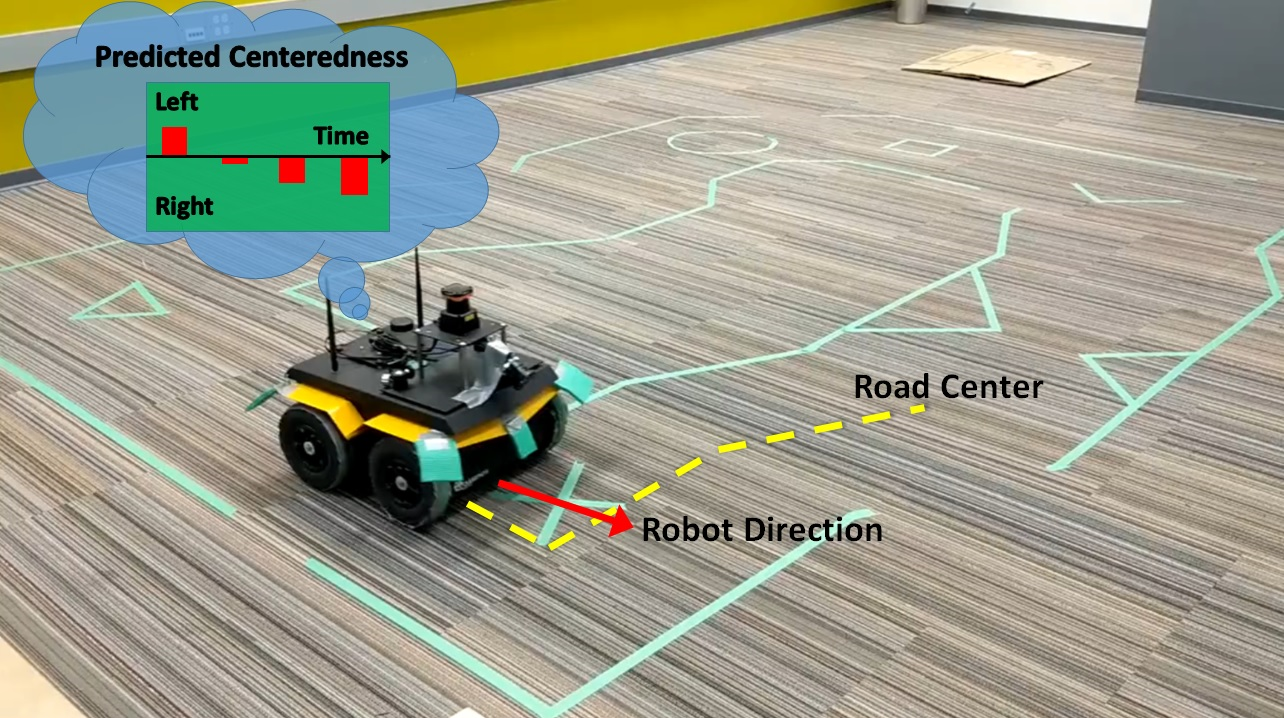}
    \caption{Jackal robot driving on a visually challenging road using GVFs as a predictive representation of state}
    \label{fig_jackal_robot_visual_steering}
\end{figure}
\begin{figure}
    \centering
    \includegraphics[width=0.45\textwidth]{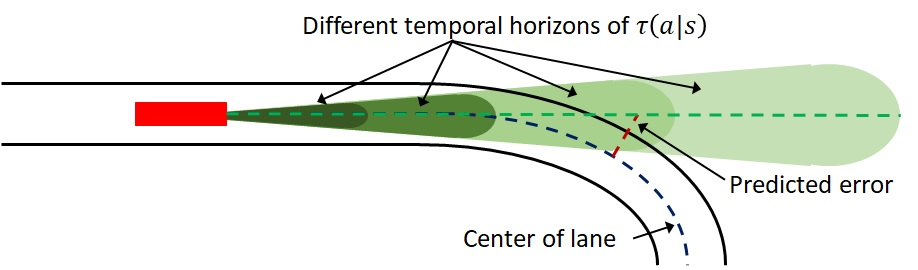}
    \caption{Depiction of the different temporal horizons of the GVF predictions for driving with GVFs}
    \label{fig_visual_steering_temporal_horizon}
\end{figure}
The lane centeredness and road offset predictions can be viewed as affordances under the policy of driving with approximately constant steering and speed actions until an out of lane event occurs.

The benefits reported include better learning, smoother control and better generalization to unseen roads \citep{graves2020}.
These advantages make for safe, fast, and comfortable autonomous driving.

\subsection{Affordance Applications in Light of GVFs}
Despite the growing number of applications of GVFs that can be related to affordances, research into the potential applications of GVFs is still in its infancy.
Our objective in this section is to claim that several successful applications of affordances in recent years, unbeknownst to the affordance community, are in fact GVF predictions and can be formalized under our proposed computational model.

We take as our first example the affordance-based solution used by \citet{zeng2019amazon} to achieve impressive results in the Amazon picking challenge.
Their solution learns to predict the success of picking up parts in a bin under different orientations, locations and motion primitives, resulting in an affordance map consisting of many thousands of predictions of success or fail.
While the authors do not relate their methods specifically to GVFs, we claim they fall under the broad and very general framework of the GVF predictions that we propose.
The cumulant is the success/fail signal, the policies are the motion primitives and termination is reached when the success or fail signal is received after executing the policy.
\citet{zeng2019amazon} learn the predictions conditioned on motion primitives using supervised learning. This approach is usually referred to as Monte Carlo estimation of value in the RL literature \citep{sutton2018}.
It has long been known to the RL community that both temporal difference (TD) learning and Monte Carlo estimation (i.e. supervised learning) can be applied to learn value predictions \citep{sutton2018}.
A more detailed analysis of the differences are provided in the Appendix.

In \citet{ugur2007traversability}, traversability predictions of regions surrounding a robot are learned to help a robot navigate through challenging environments.
The traversability predictions form a map of success or failure for different locations under possible actions of straight and various degrees of left and right turning.
They are learned using supervised learning with success or failure in traversing the terrain at a specific location being the training signal.
Like with \citet{zeng2019amazon}, we claim the traversability map prediction is a map of GVF predictions, each conditioned under different policies (e.g. straight, left, right).
The cumulants are the success or failure labels.

A very recent example is about predicting a spatial action map with RL \citep{wu2020spatial}.
Unlike the previous two examples, this one is based on RL rather than supervised learning. It is thus easier to relate it to our proposal.
The authors use Q-learning to predict a spatial action map that represents the value of moving to the goal location in pixel space.
Our proposal formalizes and broadens this way of modeling affordances.

It should be noted that these three example demonstrate a key advantage of GVF predictions: scalability.
Thousands and even millions of GVF predictions are learned and evaluated simultaneously in the form of a map in each of them.
This underwrites the point in the name given by the original paper introducing GVF predictions: a \textit{Horde} \citep{sutton2011}.
Furthermore, the examples in this section demonstrate that the scalability of GVFs is practically achievable and useful in diverse real-world problems.

\section{Directions for Future Research}
Much of the importance of our proposal lies in it foregrounding some promising directions of future research towards realizing the benefits of both affordances and GVFs in real-world applications.
We comment on four lines of future research from this perspective: (1) architectures for planning and control using affordances, (2) methods of discovering affordances, (3) reformulating robotic perception, and (4) affordance-based orchestration of complex behaviors.

\subsection{Planning and Control with GVF Predictions}
Exploitation of affordances is an important problem that requires prediction, planning and action selection to work together \citep{zech2017}.
There have been only a small number of successes in applying GVFs to planning and control of robots, in part due to the lack of established frameworks for integrating such predictions into more complex systems.
We consider three approaches to using affordances as GVFs for planning and control: (1) Pavlovian control, (2) predictive representations, and (3) model-based RL.

Pavlovian control using affordances modelled as GVFs has already enjoyed some successes in real-world experiments \citep{dalrymple2020pavlovian, graves2019a}.
While this approach requires some domain knowledge to formulate rules to link affordances with control behaviors, its value lies in the flexibility available in the design of the control rules.
For example, such flexibility allows for transparent hybrid control designs where affordances are learned, potentially with deep learning, and responses are manually specified according to the affordance predictions.
Unlike end-to-end learning solutions where it is difficult to understand how control decisions are made, Pavlovian control decisions based on semantically well-defined predictions can be readily explainable and more trusted in safety critical applications such as autonomous driving.

The second approach is to use affordances modelled as GVFs as predictive representations of states in RL learning of the optimal policy.
This approach reduces the complexity of high dimensional observations, such as the vast amounts of visual-sensory information received from cameras, to low dimensional affordance perception.
It is similarity to Pavlovian control except that the control decisions are learned with RL rather than specified by expert-defined rules.
Early work on this way of using predictive representation was in a framework called predictive state presentations (PSRs) \citep{littman2002}, where PSRs were shown to be both compact and general compared to using history alone \citep{rafols2005}.
PSRs have since been superseded by GVF predictive representations \citep{schaul2013, schlegel2018}.
\citet{schaul2013} in particular demonstrated a clear advantage in improving generalization in partially observable environments with GVF forecasts when compared to more classical PSR approaches.

Continuing along this direction, we conjecture that state can be described, at least in part, by perceived affordances, i.e. the valences of its action possibilities.
If true, the advantage of this perspective is that the space of affordances is typically much smaller than the space of observations.
This presumably could allow for significantly faster learning.
In addition, the space of affordances is presumably agnostic to the specific task at hand. This allows for very general representations that transfer well between tasks and the efficient re-use of prior knowledge in RL.
Intuitively, this approach requires that the system design, especially its $S$, $A$ and $P$ parts, and the GVF design, namely the $c$, $\tau$, $\gamma)$ parts, should fit together in a way such that at the level of GVF predictions, the Markovian property is honored for downstream RL optimization.
A future direction of research is to better understand the relationship between the state of the agent and affordances and formulate principled frameworks for learning them as general representations through unsupervised exploration.

Third, model-based RL is growing in popularity because of its potential to improve the efficiency in learning \citep{sutton2008dyna,deisenroth2011pilco,wang2019mbrlbenchmarks,khetarpal2020}.
However, a common concern is how to learn and use imperfect models.
A recent idea is to learn partial models \citep{khetarpal2020} which improves learning efficiency. Because these partial models predict the future distribution of state for taking an action in a given state, however, it can be challenging to scale such a method to large state spaces.
One possible way to address this scalability issue is using a collection of affordances modelled by GVFs as partial models that supplement or even replace primitive state and action representations with higher-level abstractions about valence and option for use in model-based RL.

\subsection{Automated Discovery of Affordances}

The rise of affordances as a matter of fit between animals and humans with their environments presumably involves many evolutionary, developmental, learning and design iterations.
For robotics, we would naturally hope that such a process could be at least partly automated and thereby accelerated.
One might even consider automated discovery of affordances a holy grail of robotics.
We believe that our proposed model as summarized in Table \ref{tab:affordance} provides valuable guidance about how to approach at least partial automation of affordance discovery.

First, our proposal links affordances to the skill discovery literature in RL \citep{eysenbach2018, gregor2016, thomas2017, achiam2018}, wherein skills are usually formulated as options \citep{sutton1999} in the absence of external rewards.
These works are still in very early stages and tend to use simplistic grid-world simulation for experimental validation.
Skill discovery through learning of options is yet to be compellingly demonstrated in real-world robotics.

One pattern we find about the option discovery literature is that it primarily focuses on learning the policy of options \citep{eysenbach2018}.
More recently, learning of the termination functions \citep{harutyunyan2019, yang2020transfer} has received more attention.
But learning the initiation sets is often ignored.
We argue that learning initiation sets \citep{graves2020lispr} is important for learning affordances for their use and especially their reuse in different task environments.

Learning skills is far from the complete picture of affordance discovery.
``How do we discover the right cumulants?'' is a question we must also answer if affordance discovery is to be truly automated.
But the discovery of cumulants has not received much attention, partly because it is very challenging and partly because in most real-world applications, it is possible to manually identify the needed cumulants according to the concerns that matter most for the application, e.g. safety, comfort, lane centeredness, etc.
In fact, we have found that a useful rule of thumb is that many of the components that go into building good reward functions, make good cumulants.
This suggests that cumulant discovery has more in common with reward decomposition or reward factoring.
While real-world applications may not be hindered until the scalability of hand-engineered cumulants become a bottleneck, we do suspect that cumulant discovery will become increasingly more important as GVF models of affordances come to be more broadly used.
It may be worthwhile for us to start thinking about the possibility of extending inverse reinforcement learning \citep{ng2000inverse, abbeel04inverse}, the effectiveness of which was compellingly demonstrated in real-world helicopter control \citep{abbeel06heli}, to ``multi-object inverse reinforcement learning'' suitable for affordance discovery.

\subsection{Reformulating Robotic Perception}

Affordance as GVF also gives us a perspective from which we can rethink robotic perception in the era of the so-called ``second-wave AI'' \citep{smith2019promise} that features deep learning and reinforcement learning.

Object detection, i.e. perceptually localizing and classifying an entity, and geometric measurements, such as structure from motion and point cloud or mesh estimation, are prevalent approaches to perception in robotics.
While these approaches are likely going to be always useful and relevant, it is a question whether these alone are enough for scaling up to full real-world complexity.
On the one hand, object detection based on supervised learning of labelled data is hardly suitable for highly diverse manipulations of myriad materials and tools, such as what a chef has to do in the kitchen on a daily basis---even if manual labelling of objects in the interaction data here could be done, many of the class labels are likely going to be irrelevant to or inadequate for guiding actual manipulation.
On the other hand, while geometric measurements in the form of 3D meshes may provide a high degree of specificity, it is not suitable for function-based perceptual generalization or transfer of skills that are a daily reality in the kitchen.
By focusing on affordances, we have a better chance of striking the right balance between relevance and specificity, and thereby providing a form of perception suitable for real-world complexity.
Indeed, some recent advancements in robotic manipulation has compellingly demonstrated this point \citep{zeng2019amazon, manuelli2019kpam}.

That object qua object of manipulation is better considered a bundle of affordances, rather than a simply classifiable and localizable entity with definite geometry, is of course a point familiar to roboticists who find the notion of affordance valuable.
What is novel, we think, is that modeling affordances as GVFs opens up a new approach with good theoretical grounding to robotic perception, an approach that can scale up to real-world complexity.
Because of the way GVFs embody direct perception, particularly through deep learning implementations, learning to perceive a large number of affordances simultaneously is possible without either explicit object detection and labelling or geometric measurements.

While our new model shows how \textit{direct perception} of affordance may be realized by learning a direct mapping from observations to predicted valence \citep{gunther2016, graves2020}, it should be noted that GVFs can also use results of object detection and geometric measurements.
This can be done by simply adding the detection and measurement results as input features for the neural networks implementing the GVF mappings.
Such architectural flexibility both supports direct perception, thereby resolving a crucial theoretical point that traditional computational models often struggle to explicate \citep{zech2017}, and is capable of subsuming existing applications of affordances in robotics that historically relied on object detection and geometric measurement \citep{hassanin2018}.

Our expectation is that \textit{scalable learning of affordances as GVFs using RL methods and DL implementations} will help usher in a new paradigm of robotic perception that both subsumes and transcends current approaches.

\subsection{Behavior Orchestration with Options and GVFs}

In real life, simple behaviors are often orchestrated to accomplish complex tasks.
At dinner table, we may lean our upper body over and stretch our arm out, before grasping the salt shaker.
An outfielder often runs while keeping an eye on the flying baseball, dives to catch it before it hits the ground, and then lets oneself slide to a stop.
In driving, we often need to brake to slow down enough before we turn the steering wheel sharply.
Autonomous driving behavior decision needs to consider speed control and lane change together \citep{urmson2008boss}.
In general, how to combine multiple relevant skills for a complex task through reasoning about affordances is both challenging and important \citep{zech2017}.

The vast majority of RL research still focuses on optimizing performance for a single task rather than learning to flexibly reuse skills by combining them appropriately.
Although the options framework \citep{sutton1999} provides a way to chain together multiple skills once they are formulated as options, it does not propose guidelines about how to formulate skills as options so that they can be chained.
We believe that our perspective of affordances as option-specific GVFs opens up a promising direction for research on behavior orchestration.

\begin{figure}
    \centering
    \includegraphics{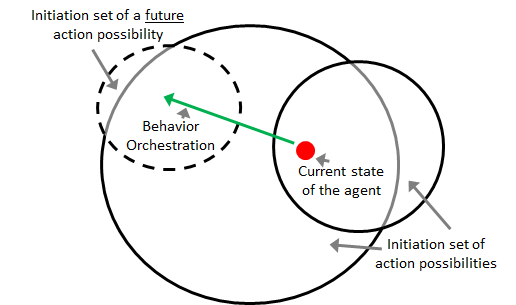}
    \caption{Two options (solid border) are currently available, with one of them capable of moving the agent into a state where a currently unavailable option (dashed border) becomes available.}
    \label{fig_options_chaining}
\end{figure}

The starting point is to take seriously the often overlooked initiation set \citep{sutton1999} of an option.
As introduced earlier, the initiation set is the set of all states where the option can be executed. 
In Figure \ref{fig_options_chaining}, there are two options, i.e. action possibilities, that are available to the agent in the current state because the agent is in the initiation set of those two options.
There is a third option that is unavailable.
However, one of the currently available options could be used to move the agent from the current state to a new state that is inside the initiation set of the third option and thereby making it available.\footnote{Naturally, the termination function of this ``bridging option'' also needs to be coordinated with the initiation set of the target option.}
This way, we can chain options together for more complex behavior.
However, because the options framework does not tell us what makes good initiation sets, models following it usually assume the initiation set to be universal.

One way to make the concept of initiation set of an option useful for behavior chaining is to take it to be the set of states where following that option leads to success.
For example, to successfully grab an object with a hand may require the arm to be in a good relative position to the object. As described earlier in the prosthesis application, GVFs can be used to predict the order of joint activation used by the amputee for moving the limb in preparation for a grasp \citep{edwards2016}.
Taking inspiration from this, if we define a cumulant which measures the success of the grasp, we can model the initiation set for the grasp option with a GVF.
The idea is that this GVF predicts the future success or failure of the option in a given state and is thus capable of representing the \textit{higher-order affordance} of the option as a single unit of action possibility suitably available or unavailable in the context.
Such a GVF may then be used to define the initiation set by thresholding for a desired minimum level of success \citep{graves2020lispr}.

The central idea of our proposal here is that by learning the initiation set of an option that is tied to its success, we may encapsulates it into an action possibility with specific valence, i.e. a proper affordance, with its specific condition of applicability captured.
We take it that this sort of modularization turns a general option into a combinable and reusable skill and enables the agent to focus on which among the currently relevant options is the most suitable in the situation at hand.

Our sketch here also raises some key challenges:
(1) How should the initiation set of an option be learned in general?\footnote{\cite{graves2020lispr} provide a first illustration, but there may be many definitions of ``success'' that depend on context. Fortunately, the generality of cumulants should be able to accommodate the required flexibility.}
(2) What are good measures of success, i.e. cumulants?
(3) How should an agent perform long-term planning using complex sequences of options as affordances?
(4) How may chaining of options support orchestration at even higher level?
Using learned initiation sets to modularize options is at best the initial impetus that we hope could catalyze much more fruitful research in this direction towards fundamentally addressing the challenge of ``behavior affording behavior'' \citep{zech2017}.

\subsection{Affordance as GVF in Future Research --- A Concrete Example}

To make our proposal more tangible, let us revisit the unprotected left turn problem illustrated in Figure \ref{fig_unprotected_left_turn} and imagine how research on it could look like from the perspective of affordance as GVF.

At any given moment, the agent needs to predict whether or not the situation affords a safe left turn through the intersection.
The \textit{policy} of the \textit{option} in question, at a high level, involves only two \textit{actions}---to \textit{proceed} with the left turn or to \textit{wait}.
The \textit{cumulant} suitable for predicting the affordance could be simply \textit{success} or \textit{failure}.
For a practical setup, we can use proximity to other vehicles, rather than absence or presence of actual collisions, to define success or failure \citep{graves2019a}.
If the ego vehicle ever gets overly close to any other vehicle as it proceeds or waits, a failure is registered; success is when the vehicle finishes the left turn without failure.
Given this setup, we can then collect data of real-world interaction trajectories at many intersections and use off-policy and off-line RL methods \citep{graves2020} to train a \textit{neural network} to predict whether it will be safe to turn left through the intersection.
Based on such a prediction of affordance, the actual decision making could be done with a simple rule: if \textit{safe} then \textit{proceed} else \textit{wait}.

Conventional approach to predicting whether it is safe for a left turn relies on accurate detection of other vehicles, precise estimation of their speed, and explicit modeling and prediction of their future motion \citep{urmson2008boss}.
In contrast, while the GVF approach can use proximity derived from detection of individual vehicles, proximity can also be acquired without explicit individuation and localization of other dynamic objects \citep{jin2020mapless}.
Moreover, because the neural network that implements the prediction can map high dimensional data directly to \textit{valence} of affordance, extending the solution to bad weather conditions and alternative driving cultures may be done through simply collecting the relevant interaction trajectories and retraining. Neither explicit labelling of bad weather data nor extra modeling of alternative driver behavior is necessary.

In supervised learning of affordance prediction, such as in \citep{zeng2019amazon}, each interaction trajectory through the whole intersection yields only a single data point according to the final outcome.
Given that real world data is going to be heavily skewed towards successful left turns, many failed attempts will have to be deliberately collected.
In contrast, under the GVF approach, every time step in the data, so long as it is tied to an appropriate cumulant, yields a meaningful training signal.
Thus, the rare failures can be much more efficiently used in learning affordance prediction.

Now, for many social and technical reasons, we may doubt that the GVF approach to affordance will soon play a key role in solving the unprotected left turn challenge in the real world.
But such a possible future is at least imaginable.

\section{Conclusion}


Understanding the importance of GVFs in real-world applications has been a long sought-after goal since their introduction \citep{sutton2011} and subsequent treatment \citep{awhite2015, white2017}.
By connecting GVFs with affordances, we hope to catalyze further research and development of real-world applications that benefit both the RL and the robotics communities.
We highlighted the relationship between GVFs and affordances in many compelling real-world applications that showcase the utility and broad applicability of GVF predictions to many real-world problems.

Our central message is that affordances can be modelled by GVFs when formalized as a combination of (1) valence predictions based on cumulants that represent some salient features about the agent-environment relationship and (2) action possibilities formulated as options.
The advantages of modelling affordances as GVFs include (1) realizing affordance prediction as a form of direction perception when combined with deep learning, (2) concretizing the fundamental connection between action and perception that is the hallmark of affordances, and (3) offering a scalable way for learning affordances.

In summary, modelling affordances as GVFs allows the fields of psychology and robotics to benefit from the field of RL and raises deep challenges about how to support planning and control in complex real-world applications using affordances, how to discover or initially learn affordances, how to do robotic perception that is adequate for complex real-world interactions, and how to orchestrate behaviors that afford other behaviors.
We believe that further developing and exploiting this important link between GVFs and affordances will deepen interdisciplinary collaboration towards developing more intelligent machines with more general capabilities for more complex behaviors. By systematically making this link explicit, we hope that both GVFs and affordances will receive the attention they deserve in robotics and artificial intelligence.

\bibliographystyle{SageH}
\bibliography{references}

\clearpage
\appendix
\section{Appendix}
\subsection{Algorithms for learning GVFs}
GVFs can be learned with temporal difference learning where a prediction of value at a future state--- usually the next state in one-step learning---is used to update the value at the current state.
Formally, the GVF $V^{\tau}(s)$ in state $s$ is given by the recursive relationship

\begin{equation}
V^{\tau}(s)=\E_{\tau}[c_{t+1}+V^{\tau}(s_{t+1})|s_t=s]
\label{eq_td_value}
\end{equation}

\noindent
where the notation $\E_{\tau}$ is the expectation over the policy $\tau$ and its state visitation distribution.
The usual way to learn $V^{\tau}(s)$ is to parameterize it as $V^{\tau}(s; \theta)$ and then learn $\theta$ by exploiting the recursive relationship in Equation \ref{eq_td_value} to compute the so-called \textit{temporal difference (TD) error} $\delta_t$ given by the following equations:

\begin{equation}
\delta_t=V^{\tau}(s_t;\theta_t)-y_t
\label{eq_td_error}
\end{equation}
\begin{equation}
y_t=c_{t+1}+V^{\tau}(s_{t+1};\theta_t)
\label{eq_td_target}
\end{equation}

\noindent
where $y_t$ is the \textit{target}.

The loss function for learning the GVF is given by

\begin{equation}
L(\theta)=\E_{\tau}[\delta^2]=\E_{\tau}[(V^{\tau}(s;\theta)-y)^2].
\label{eq_td_loss}
\end{equation}

\noindent
Most approaches minimize the loss $L(\theta)$ by taking the derivative with respect to $\theta$ and updating the parameters with this gradient.
However, it is worth pointing out that $y_t$ depends on $\theta$, given Equation \ref{eq_td_target}. In practice, this dependency is often ignored by taking $y_t$ to be a fixed target independent of $\theta$ when computing the gradient.
Residual TD methods provide an alternative approach to address this bias, but it is considerably more complicated to implement and not well understood in deep reinforcement learning \citep{ghiassian2020}.
Therefore, the most common approach is to apply the ``on-policy'' update with the gradient

\begin{equation}
\nabla_{\theta}L(\theta)=\E_{\tau}[\delta \nabla_{\theta}V^{\tau}(s;\theta)]
\label{eq_td_gradient}
\end{equation}

\noindent
which assumes that the samples used to calculate the update are collected with the policy $\tau$.

In situations where a different behavior policy $\mu$ is used to collect the data, the update is said to be ``off-policy''.
In this case, one must correct for the change in the the behavior policy distribution $\mu(a|s)$ from the desired distribution $\tau(a|s)$ \citep{radiee2019}.
We will simplify the notation and refer to $\mu$ and $\tau$ as policy distributions over action conditioned on state.
The loss gradient is thus given by 

\begin{equation}
\nabla_{\theta}L(\theta)=\E_{\mu}[\rho \delta \nabla_{\theta}V^{\tau}(s;\theta)]
\label{eq_td_gradient_offpolicy}
\end{equation}

\noindent
where $\rho=\frac{\tau}{\mu}$ is the importance sampling ratio that corrects for the difference.
Note that $\rho$ does not correct for the state visitation distribution since the expectation is still under the behavior policy $\mu$; however, in practice this is not an important issue \citep{schlegel2019}.
The importance sampling ratio can result in high variance updates though.
Fortunately, there is an alternative approach using importance resampling that is shown to give lower variance updates where the correction is applied by sampling mini-batches from a buffer of experience proportional to the estimated importance sampling ratio $\hat{\rho} = \frac{\rho}{\sum{\rho}}$ \citep{schlegel2019}.
To eliminate the bias in using a finite replay buffer of experience to estimate the importance sampling ratio $\hat{\rho}$, a correction factor $\bar{\rho}$ is needed which is the average $\rho$ in the buffer of experience:

\begin{equation}
\nabla_{\theta}L(\theta)=\E_{\mu, \hat{\rho}}[\bar{\rho} \delta \nabla_{\theta}V^{\tau}(s;\theta)]
\label{eq_td_gradient_offpolicy_resampling}
\end{equation}

\noindent
The expectation is under the behavior policy distribution $\mu$ where the transitions beginning with state $s$ are sampled from the buffer of experience proportional to the estimated importance sampling ratio $\hat{rho}$.
In practice, if the buffer of experience is large enough, the correction factor $\bar{\rho}$ is not necessary.

A similar recursive relationship exists for GAVF $Q^{\tau}(s,a)$.

\begin{equation}
\begin{split}
Q^{\tau}(s,a)= & \E_{\tau}[c_{t+1}+Q^{\tau}(s_{t+1},a') \\
& | s_t=s, a_t=a, a' \sim \tau]
\end{split}
\label{eq_td_action_value}
\end{equation}

\noindent
The TD error is given by

\begin{equation}
\delta_t=Q^{\tau}(s_t,a_t;\theta_t)-y_t
\label{eq_td_action_error}
\end{equation}
\begin{equation}
y_t=c_{t+1}+\E_{a' \sim \tau}[Q^{\tau}(s_{t+1},a';\theta_t)]
\label{eq_td_action_target}
\end{equation}

The loss function for learning the GAVF is given by 

\begin{equation}
L(\theta)=\E_{\tau}[\delta^2]=\E_{\tau}[(Q^{\tau}(s,a;\theta)-y)^2]
\label{eq_gavf_loss}
\end{equation}

\noindent
and the loss gradient is
\begin{equation}
\nabla_{\theta}L(\theta)=\E_{a \sim \mu}[\delta \nabla_{\theta}Q^{\tau}(s,a;\theta)].
\label{eq_td_action_gradient}
\end{equation}

It is worth noting that the on-policy and off-policy updates for GAVFs are the same, because the action in the bootstrapped prediction $y_t$ (Equation \ref{eq_td_action_target}) is an expectation over the policy $\tau$ eliminating the need for a correction between the mismatching behavior distribution $\mu$ and $\tau$.
Thus, in settings where the behavior policy $\mu$ is different from $\tau$, a GAVF can be easier to learn.
The GAVF is also more general and can be used to to obtain the GVF in practice according to $V^{\tau}(s)=\E_{a \sim \tau}[Q^{\tau}(s,a)]$.

The algorithm for learning the GVF predictions off-policy is summarized in Algorithm \ref{alg_gvf_offpolicy_train} where the continuation function $\gamma$ introduced in the main text is also considered.
For cases where the behavior policy $\mu$ is unknown, \cite{graves2020} introduced a method to estimate it.
This method can be adapted to learning GVFs offline from batch data, including data collected from human agents \citep{graves2020}.

\begin{algorithm}
\caption{Off-policy GVF learning algorithm}
\label{alg_gvf_offpolicy_train}
\begin{algorithmic}[1]
\State Initialize $\hat{v}^\tau$, and replay memory $D$, 
\State Observe initial state $s_0$
\For{t=0,T}
  \State Sample action $a_t$ from $\mu(a_t|s_t)$
  \State Execute action $a_t$ and observe state $s_{t+1}$
  \State Compute cumulant $c_{t+1}=c(s_t,a_t,s_{t+1})$
  \State Compute continuation $\gamma_{t+1}=\gamma(s_t,a_t,s_{t+1})$
  \State Compute importance sampling ratio $\rho_t=\frac{\tau(a_t|s_t)}{\mu(a_t|s_t)}$
  \State Store transition $(s_t,a_t,c_{t+1},\gamma_{t+1},s_{t+1},\rho_t)$ in $D$
  \State Compute average importance sampling ratio in replay buffer $\bar{\rho}=\frac{\sum_{j=1}^{n}{\rho_j}}{n}$
  \State Sample transitions $(s_i,a_i,c_{i+1},\gamma_{i+1},s_{i+1})$ from $D$ according to probability $\frac{\rho_i}{\sum_{j=1}^{n}\rho_j}$ to form a minibatch $A$
  \State Compute $y_i = c_{i+1} + \gamma_{i+1} \hat{v}^\tau(s_{i+1};\theta)$ for minibatch $A$
  \State Perform gradient descent step on $(y_i-\hat{v}^\tau(s_i;\theta))^2$ according to $\nabla_{\theta}L(\theta)=\bar{\rho} \delta \nabla_{\theta}V^{\tau}(s;\theta)$ for minibatch $A$
\EndFor
\end{algorithmic}
\end{algorithm}

\subsection{GVFs versus Other Prediction Methods} \label{app:gvf-comparison}
There are predominantly two ways to predict the future: (1) predict the sum of future cumulants \citep{sutton2011}, (2) predict a sequence of states using a state transition model \citep{sutton1988}, (3) predict the final outcome \citep{zeng2019amazon, ugur2007traversability, wu2020spatial}.
The latter is perhaps the  most common approach found in affordance prediction literature.
Our goal in this section is to show how all three of these predictions can be framed as GVF predictions.

\subsubsection{GVFs and Supervised Affordance Predictions}
It has long been recognized that temporal difference learning and supervised learning can be used to solve the same prediction problem \citep{sutton1988}.
With supervised learning, the learner is provided with final outcomes rather than intermediate cumulants.
We will demonstrate that GVFs are flexible enough to include the conventional supervised affordance prediction as a special case.

Consider the conventional affordance prediction where an agent is in state $s_t$ when it executes some action possibility leading to an outcome label $c_T$ in state $s_T$ for some time step in the future $T>t$ where the outcome label $c_T$ indicates the success or failure of the action possibility.
The conventional affordance prediction of the final outcome of an action possibility can be framed as a GVF prediction under the following conditions:

\begin{itemize}
\item Use $\gamma=1.0$ for undiscounted predictions, and
\item Use cumulant $c_T$ for states $s_T$ with final outcomes and $c_t=0$ otherwise.
\end{itemize}

Formally, the conventional affordance prediction that terminates in state $s_T$ is given by

\begin{equation}
    \begin{split}
        V^{\tau}(s_t) & = \E_{\tau}[\sum_{i=0}^{T-1}{\gamma^i c_{t+i+1}}]\\
            & = \E_{\tau}[\sum_{i=0}^{T-1}{c_{t+i+1}}]\\
            & = \E_{\tau}[c_T]
    \end{split}
\end{equation}

The value prediction is just a prediction of the final outcome which can be learned by supervised learning.
Thus, the conventional affordance prediction can be cast as a GVF learning problem.

GVFs offers additional flexibility over the conventional (supervised) affordance prediction in that they allow intermediate cumulants (before final outcome) that can be used to evaluate how the agent arrived at the final outcome rather than a simple success/fail for reaching the desired state.
In addition, learning GVFs off-policy where $\tau$ is different from the policy used to collect the data is very challenging with supervised learning since the importance sampling ratios at each time step are multiplied together causing many trajectory probabilities to vanish over long horizons.
Temporal different learning is better suited for off-policy corrections since the importance sampling ratio is only applied for the immediate transition and not the entire sequence.

\subsubsection{GVFs and Next-Step Transition Models}
Formally, a state transition model is a distribution over next states given as $p(s_{t+1}|a_t, s_t)$.
It is well known that predicting future states across long time horizons results in accumulation of errors with this kind of approach \citep{sutton1988}.
To see how badly errors in the transition model can accumulate, let us consider asking the question, what is the probability of observing the trajectory (i.e. state-action sequence) ending in state $s_T$ at some time $T > t$ for some starting state $s_t$ under the policy $\pi(a|s)$.
This simplifies to
\begin{equation}
    p(s_T,a_{T-1},s_{T-1}, ... | s_t) = \prod_{k=t}^{T-1}p(s_{k+1}|s_{k},a_{k})\pi(a_{k}|s_{k})
\end{equation}

\noindent
It is easy to see how even small errors in the transition model $p(s_{t+1}|a_t, s_t)$ lead to multiplicative errors in the estimation of $p(s_T,a_{T-1},s_{T-1}, ... | s_t)$ \citep{hautsch2012mem}.

In contrast, GVFs predict summaries of future trajectories under a given policy.
However, GVFs are very general because they include next step models as a special case under certain conditions, namely:

\begin{itemize}
\item Set $\gamma=0.0$ for myopic prediction, and
\item Use a vector of cumulants that exactly corresponds to the state vector, i.e. $c_t=s_t$.
\end{itemize}

\noindent
Formally, the GVF under these conditions reduces to
\begin{equation}
    \begin{split}
        V^{\tau}(s_t) & = \E_{\tau}[\sum_{i=0}^{\infty}{\gamma^i c_{t+i+1}}]\\
            & = \E_{\tau}[s_{t+1}]
    \end{split}
\end{equation}

\noindent
This GVF prediction is under the policy $\tau$.
However, note that an action-conditioned model learned with GAVFs under these conditions reduces to an action-conditioned next state model prediction and does not require the policy $\tau$.

What we hope to illustrate here two points: (1) next-step models can be framed as GVFs, and (2) the advantage of GVFs is greater flexibility dictated by the choice of $\gamma$ and the cumulant.
Additionally, next state models with infinite horizon predictions of $V^{\tau}(s)$ are not computationally tractable and thus approximations with finite horizon predictions are normally used.
GVFs avoid (1) computational burden of predicting all future states, (2) approximation error with finite horizon predictions, and (3) multiplicative error from iteration with the model.
We summarize a comparison of the RL value predictions and conventional next-step predictions in Table \ref{tab:prediction_next_step_comparison}.

\end{document}